# Iterative exact global histogram specification and SSIM gradient ascent: a proof of convergence, step size and parameter selection

Alireza Avanaki

user@yahoo.com (my ID is my last name)

**Abstract**

The SSIM-optimized exact global histogram specification (EGHS) is shown to converge in the sense that the first order approximation of the result's quality (i.e., its structural similarity with input) does not decrease in an iteration, when the step size is small. Each iteration is composed of SSIM gradient ascent and basic EGHS with the specified target histogram. Selection of step size and other parameters is also discussed.

## I. Introduction and background

See [1] or [2] for the details of exact global histogram specification (EGHS) optimized for structural similarity (SSIM) [3].

## II. Convergence analysis

We need the following lemma first. See [1] or [2] for notations, Algorithm 1, and basic (a.k.a. classic) EGHS.

**Lemma.** If the histogram of $Y$ is $H$, and

$$X = Y + \beta \nabla_Y \text{SSIM}(I,Y), \quad \text{with } \beta > 0 \tag{2}$$

$$Y' = \text{EGHS}(X, H), \tag{3}$$

then the histogram of $Y'$ is also $H$, and the first order approximation of $\mathrm{SSIM}(I,Y')$ is not less than $\mathrm{SSIM}(I,Y)$. $\nabla_Y$ denotes gradient with respect to image $Y$ (i.e., $\nabla_Y = \left(\frac{\partial}{\partial y_i}\right)_{i=1,\ldots,M}$, where $y_i$ is the value of the $i^{\text{th}}$ pixel of $Y$), and $\mathrm{EGHS}(.,.)$ denotes the basic EGHS method (i.e., Algorithm 1, with no auxiliary sorting information).

**Proof.** The histogram of $Y'$ is $H$ because of (3). We show that the first order change in SSIM introduced by (2) and (3) is non-negative. Consider a pixel of $Y$, with intensity $y$ (i.e., in bin $y$ of $H$) and updated intensity $x$, given by (2), that is assigned to bin $y'$ by (3). The following six possible cases can be distinguished: (i) $y \leq y' \leq x$, (ii) $y \leq x \leq y'$, (iii) $x \leq y \leq y'$, (iv) $x \leq y' \leq y$, (v) $y' \leq x \leq y$, and (vi) $y' \leq y \leq x$.

In case (i) the overall change in $y$, given by $y' - y$, has the same sign as the gradient, given by $\beta^{-1}(x-y)$. Hence, the change in SSIM contributed by this pixel, given by $\beta^{-1}(x-y)(y'-y)$, is non-negative. A similar argument also holds for cases (ii), (iv), and (v).

In case (vi), if $y' = y$ or $x = y$, the change in SSIM contributed by this pixel is zero. Therefore we consider $y' < y < x$. EGHS does not push back $x$ to fill a vacancy on bin $y'$, unless there is a pixel from bin $y_0$, with $y_0 \leq y'$, for which $x_0 \geq x$. If $y'_0 \geq y$, the change in SSIM by these two pixels is given by

$$\Delta\mathrm{SSIM} = \beta^{-1}\left[(x_0 - y_0)(y'_0 - y_0) - (x-y)(y-y')\right]$$
$$\geq \beta^{-1}\left[(x_0 - y_0)(y - y_0) - (x-y)(y-y')\right] \qquad (4)$$

From the assumptions, we have $x_0 - y_0 \geq x - y$, which yields $\Delta\mathrm{SSIM} \geq \beta^{-1}(x-y)(y'-y_0) \geq 0$.

If $y'_0 < y$, there must be a pixel $y_1$, with $y_1 < y$, for which $x_1 \geq x_0$, or EGHS does not push back $x_0$ to fill a vacancy on bin $y'_0$. If $y'_1 \geq y$, we have

$$\Delta\text{SSIM} = \beta^{-1}[(x_1 - y_1)(y'_1 - y_1) + (x_0 - y_0)(y'_0 - y_0) - (x - y)(y - y')] \\ \geq \beta^{-1}[(x_1 - y_1)(y - y_1) + (x_0 - y_0)(y'_0 - y_0) - (x - y)(y - y')] \tag{5}$$

From the assumptions, we have $(y - y_1) + (y'_0 - y_0) \geq y - y'$, which yields

$$\Delta\text{SSIM} \geq \beta^{-1}[(x_1 - y_1 - x + y)(y - y_1) + (x_0 - y_0 - x + y)(y'_0 - y_0)], \tag{6}$$

which is non-negative because $x_1 \geq x$, $y > y_1$, $x_0 \geq x$, $y > y_0$, and $y'_0 \geq y_0$.

If $y'_1 < y$, we can go on by studying the effect of another vacancy under $y$ that caused $y'_1 < y$, and get a non-negative lower bound on the overall change on SSIM. Eventually, we will have $y'_n \geq y$ as the number of possible vacancies under $y$ is limited. End of proof for case (vi).

Case (iii) is the same as case (vi) with directions of the gradient and the change in $y$ reversed. Hence the proof of this case is similar to case (vi).

Q.E.D.

This lemma guarantees that the first order approximation of the EGHS solution quality (i.e., SSIM(*I*, *Y*)) increases for $\beta > 0$ by the following algorithm (except in special cases discussed below).

**Algorithm 2. SSIM-enhanced EGHS solution**

    Input image $I$ and target histogram $H$
    $Y = \text{EGHS}(I, H)$
    While 1,
        $X = Y + \beta \nabla_Y \text{SSIM}(I, Y)$
        $Y = \text{EGHS}(X, H)$
        If stopping crietrion is met, break
    End while
    Output $Y$

Algorithm 2 is a gradient ascent in the subspace of images with histogram $H$. The first line in the loop enhances the quality of $Y$, an EGHS solution. Such enhancement changes the histogram. The second line projects the enhanced solution to a valid EGHS solution, with a quality better (or the same as) that of the EGHS solution from the last iteration. The stopping criterion may be either (or a combination) of the following: (i) The solution quality is above a given threshold; (ii) The growth in solution quality is under a certain threshold. (iii) The number of iterations reaches a limit. The latter is used in the experiments of Section III.

In the following cases the solution quality is not increased in the loop. (i) The step size, $\beta$, is too small. In this case, EGHS "quantizes" $Y + \beta \nabla_Y \text{SSIM}(I, Y)$ back to $Y$. Hence the solution quality remains constant; (ii) the step size is too large. In this case $Y + \beta \nabla_Y \text{SSIM}(I, Y)$ advances too far in the gradient direction and skips over the maximum. Thus, $\text{SSIM}(I, X)$ can be smaller than $\text{SSIM}(I, Y)$; (iii) $\|\nabla_Y \text{SSIM}(I, Y)\|$ vanishes. Then $Y = I$, since the histogram of the input is already $H$.

### III. Step size selection

Using a step size that is too large, we may skip over the maximum during gradient ascent. A step size that is too small, on the other hand, requires a lot more iterations for a certain increase in SSIM. In the following, we derive the step size that yields the maximum SSIM growth in each iteration. Although this greedy approach may not necessarily lead us to *the* highest quality EGHS solution, it enhanced the quality of the result in all of our experiments.

The value of step size that maximizes SSIM growth in each iteration is given by:

$$\beta_{opt} = \arg\max_{\beta} \text{SSIM}(I, \text{EGHS}(Y + \beta \nabla_Y \text{SSIM}(I,Y), H)) \tag{7}$$

This maximization problem is difficult because EGHS is not differentiable. We observed that EGHS behavior can be modeled by a gain less than unity on SSIM of its image argument for values of $\beta$ about (within the bounds described below) $\beta_{opt}$. Thus, we can approximate $\beta_{opt}$ by

$$\beta_{opt} = \arg\max_{\beta} \text{SSIM}(I, Y + \beta \nabla_Y \text{SSIM}(I,Y)). \tag{8}$$

To solve this, we set

$$\frac{\partial}{\partial \beta} \text{SSIM}(I, Y + \beta \nabla_Y \text{SSIM}(I,Y)) = 0. \tag{9}$$

By application of the chain rule and substitution of the first order approximation of $\nabla_Y \text{SSIM}(I,X)$ we get

$$\nabla_Y \text{SSIM}(I,Y) \cdot \left( \nabla_Y \text{SSIM}(I,Y) + \beta H_Y \text{SSIM}(I,Y) \nabla_Y \text{SSIM}(I,Y) \right) = 0, \tag{10}$$

in which $H_Y = \left( \frac{\partial^2}{\partial y_i \partial y_j} \right)_{i,j=1,\ldots,M}$ is the hessian operator, and '.' denotes the dot product. Hence,

$$\beta_{opt} = \frac{-\|\nabla_Y \text{SSIM}(I,Y)\|^2}{\left(\nabla_Y \text{SSIM}(I,Y)\right)^T H_Y \text{SSIM}(I,Y) \nabla_Y \text{SSIM}(I,Y)}, \tag{11}$$

where $^T$ denotes vector transposition ($\nabla_Y$ is $M \times 1$ and $H_Y$ is $M \times M$) and $\|V\|^2 = V \cdot V = V^T V$.

Since the calculation of SSIM hessian, required in (11), is cumbersome, we also compute an upper bound for $\beta_{opt}$ in the following. First note that the increase in SSIM by (2) is given by $\beta \|\nabla_Y \text{SSIM}(I,Y)\|^2$ which must be less than $1 - \text{SSIM}(I,Y)$, since the maximum possible value of $\text{SSIM}(I,X)$ is one. Thus, we have

$$\beta_{opt} \leq \frac{1 - \text{SSIM}(I,Y)}{\|\nabla_Y \text{SSIM}(I,Y)\|^2} . \qquad (12)$$

To get a lower bound on the step size, note that if $\max|\beta \nabla_Y \text{SSIM}(I,Y)| < 0.5$, no pixels crosses the boundary between consequent histogram bins (apart by 1). In other words, $\beta \nabla_Y \text{SSIM}(I,Y)$ remains within the "dead-zone" of EGHS, hence EGHS(X, H) becomes exactly Y. Therefore, to enhance SSIM iteratively we need $\beta \max|\nabla_Y \text{SSIM}(I,Y)| \geq 0.5$ which translates to

$$\beta_{opt} \geq \frac{1}{2 \max|\nabla_Y \text{SSIM}(I,Y)|} . \qquad (13)$$

Instead of using (11), the optimal value of step size can be found with a scalar search between the bounds given by (12) and (13).

### IV. SSIM parameter selection

The low-pass kernel $W$ used in computation of SSIM and its gradient reduces the effect of high-frequency components (i.e., very small details) of the input images in measuring their similarity. Wang *et al.* suggests a Gaussian kernel with standard deviation of 1.5 (truncated to 11x11 & normalized) so that SSIM conforms best to perceptual quality [3].

The suggested values of the other SSIM parameters are $C_1 = (K_1 L)^2$ and $C_2 = (K_2 L)^2$, where $L$ is the number of possible intensity levels (e.g., 256 for 8-bit images), and $K_1 = 0.01$, and $K_2 = 0.03$; although the SSIM performance is found to be "fairly insensitive to variations of these values" [3]. That is while in our experiments, we found variations of $K_1$ and $K_2$ affect the result quality.

To see the effect of $C_1$, let us inspect its relevant terms in SSIM map: $\frac{2\mu_X \mu_Y + C_1}{\mu_X^2 + \mu_Y^2 + C_1}$. In this expression, $C_1$ limits the impact of the dark areas (with low average intensity) on SSIM map. Therefore, by decreasing $C_1$, the dark areas are involved with a higher weight in determination of SSIM. Similarly by inspection of $\frac{2\sigma_{XY} + C_2}{\sigma_X^2 + \sigma_Y^2 + C_2}$ we see when $C_2$ is decreased, the low energy (smooth) areas matter more in SSIM calculation.

We observed that by using values of $K_1$ and $K_2$ suggested in [3], the areas of the input that are smooth *and* dark suffer considerable loss of details in the result of the proposed method. Empirically, we found that the values of $K_1$ and $K_2 = K_1$ between 0.003 and 0.005 strike a balance between stability of the algorithm and preservation of details in the result.